\documentclass[runningheads]{llncs}
\usepackage{graphicx}

\usepackage{tikz}
\usepackage{comment}
\usepackage{amsmath,amssymb} 
\usepackage{color}

\usepackage[accsupp]{axessibility}  

\begin{document}
\pagestyle{headings}
\mainmatter
\def\ECCVSubNumber{12}  

\title{On Biased Behavior of GANs for Face Verification} 

\titlerunning{Biased Behavior of GANs}
%

\author{Sasikanth Kotti \and
	Mayank Vatsa \and Richa Singh  \\
	{ \{kotti.1,mvatsa,richa\}@iitj.ac.in}}
\authorrunning{Sasikanth et al.}
%
\institute{IIT Jodhpur, India}

\maketitle

\begin{abstract}
	Deep Learning systems need large data for training.
	Datasets for training face verification systems are difficult to obtain and
	prone to privacy issues. Synthetic data generated by generative models
	such as GANs can be a good alternative. However, we show that data
	generated from GANs are prone to bias and fairness issues. Specifically, 
	GANs trained on FFHQ dataset show biased behavior towards generating white
	faces in the age group of 20-29. We also demonstrate that synthetic faces
	cause disparate impact, specifically for race attribute, when used for fine
	tuning face verification systems. 
\end{abstract}

\section{Introduction}

Generative Models such as Generative Adversarial Networks (GANs)~\cite{goodfellow2014generative,zhu2017unpaired,arjovsky2017wasserstein} are basic building blocks in most of image recognition architectures. The task of face verification ~\cite{chopra2005learning,huang2012learning,sun2013hybrid,lu2015surpassing} consists of verifying if the given pair of faces belongs to the same identity. Deep Learning based algorithms for face recognition ~\cite{turk1991face,belhumeur1997eigenfaces,an2022killing,liu2022learning} and verification ~\cite{chopra2005learning,huang2012learning,sun2013hybrid,lu2015surpassing} utilize face datasets for training. However, obtaining more data is not always easy and even sometimes not possible. GANs can be used to obtain synthetic data where data is scarce and in scenarios where privacy is important. However, existing models (GANs) trained with FFHQ dataset ~\cite{Karras_2019_CVPR} are prone to bias and fairness issues. In this work, we analyze bias and fairness of GANs ~\cite{lang2021explaining} and their impact on face verification systems. Our main contributions in this research are as follows :

\let\labelitemi\labelitemii
\begin{itemize}
	\item \textbf{Result-1:} We observed that GANs trained on FFHQ dataset exhibit bias for the "age" and "race" protected attributes.
	\item \textbf{Result-2:} We demostrate that Face Verification systems that are trained or fine-tuned with GAN data exacerbate bias for the "race" protected attribute.
\end{itemize}

\section{Datasets and Protocol}

In this section we briefly describe the datasets and evaluation protocol. In the experiments, we use three datasets: 

\begin{itemize}
	\item The Balanced Faces in the Wild (BFW)~\cite{robinson2020face} dataset is balanced across eight subgroups. This consists of 800 face images of 100 subjects, each with 25 face samples. The BFW dataset is grouped into ethnicities (i.e., Asian (A), Black (B), Indian (I), White (W)) and genders (i.e., Females (F) and Males (M)).
	
	\item CMU Multi-PIE~\cite{gross2010multi} is a constrained dataset consisting of face images of 337 subjects with variation in pose, illumination and expressions. Of these over 44K images of 336 subjects images are selected corresponding to frontal face images having illumination and expression variations.
	
	\item FFHQ or Flickr-Faces-HQ~\cite{Karras_2019_CVPR} is a dataset of 70,000 human faces of high resolution 1024x1024 and covers considerable diversity and variation.
\end{itemize}

Evaluation for estimation of bias and fairness is performed in two phases. Initially, the proportion of faces generated for each sub-group of different attributes such as Age, Gender, Race and Race4 were analysed. In the next phase, a pretrained face verification model is fine-tuned, and the impact of fairness is analyzed using Degree Of Bias(DoB) metric. We define DoB for face verification as the standard deviation of GAR@FAR

\begin{equation}
	DoB_{fv} = \sqrt\frac{\sum{(GAR_{sg}-\mu)^2}}{N}
\end{equation}  

\noindent where $GAR_{sg}$ stands for GAR @ FAR for each sub-group, $\mu$ represents mean of GAR@FAR and $N$ represents number of sub-groups. The GAR and FAR stands for Genuine Accept Rate and False Accept Rate respectively.

\section{Experiments}

\noindent  \textbf{Experiment-1:} As part of this experiment the generator of StyleGAN2 with adaptive discriminator augmentation (ADA)~\cite{Karras2020ada} trained on the FFHQ dataset is used to generate synthetic face images. Attributes such as race, race4, gender and age of these generated synthetic faces were obtained using a pretrained Fairface ~\cite{karkkainenfairface} attribute classifier. The proportion of images for each attribute type were plotted, to understand bias and imbalance.

\noindent  \textbf{Experiment-2:} In this experiment DiscoFaceGAN ~\cite{Deng_2020_CVPR} is considered for generating different faces for different identities, expressions, lighting and poses.
VGGFace2 ~\cite{cao2018vggface2} model is considered for domain adaptation with CMU Multi-PIE ~\cite{gross2010multi} and synthetic faces generated with DiscoFaceGAN ~\cite{Deng_2020_CVPR}. 

About 10000 synthetic faces of 2500 identities were generated with DiscoFaceGAN ~\cite{Deng_2020_CVPR}. Out of these, 2000 identities were used for training and, 500 identities were used for validation. Similarly the 336 subjects of CMU Multi-PIE ~\cite{gross2010multi} were split into 70-30 ratio for training and validation. Fine-tuning with both datasets was carried out for 10 epochs with a learning rate of 1e-4 , batch size of 128, weight decay of 1e-4 and momentum of 0.9. The last two convolutional layers of VGGFace2 [6] were fine-tuned with ArcFace ~\cite{deng2019arcface} loss of margin 35 and scale 64. The checkpoint with the lowest validation loss for each dataset is considered for inference with BFW dataset~\cite{robinson2020face}.

Inference is carried out by using Cosine distance between the pairs. Comparison of $DoB_{fv}$ i.e Std(GAR @ FAR) for different attributes such as race, gender and others is carried out for both models.

\begin{figure}[!t]
	\centering
	\includegraphics[width=\textwidth, height=4.5cm]{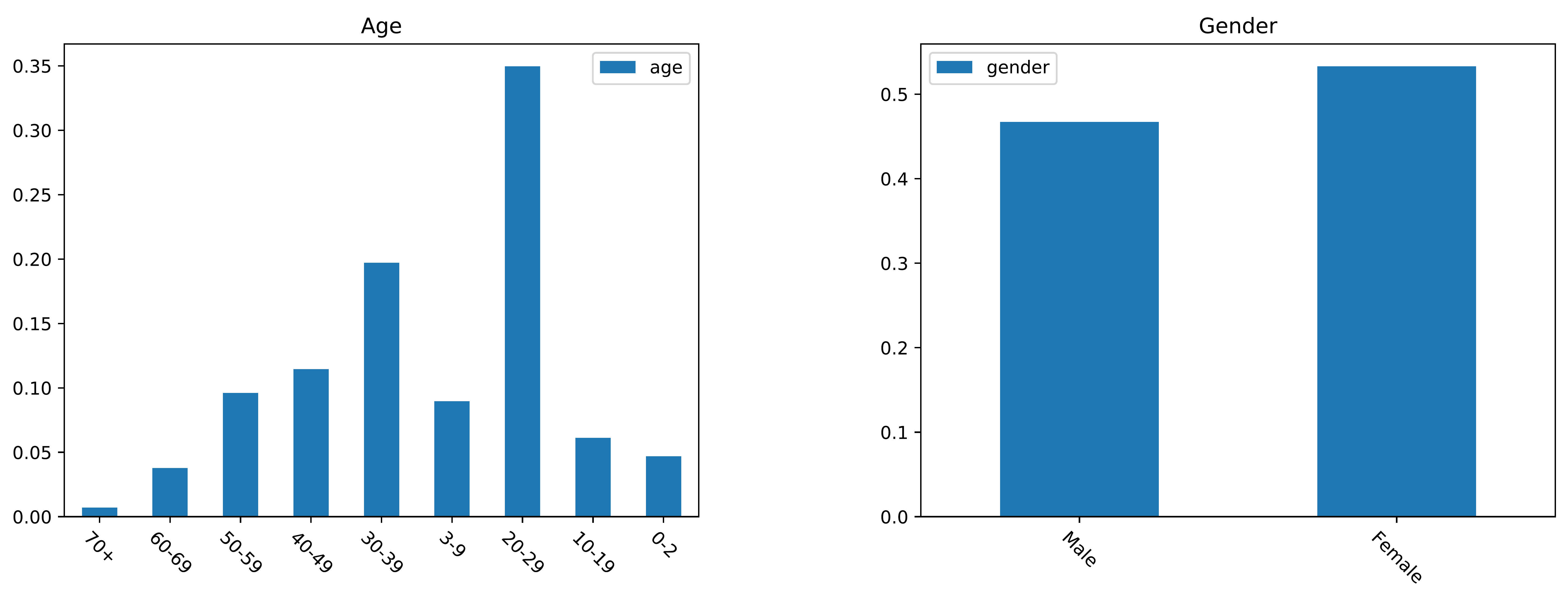}
	\centering
	\includegraphics[width=\textwidth,height=4.5cm]{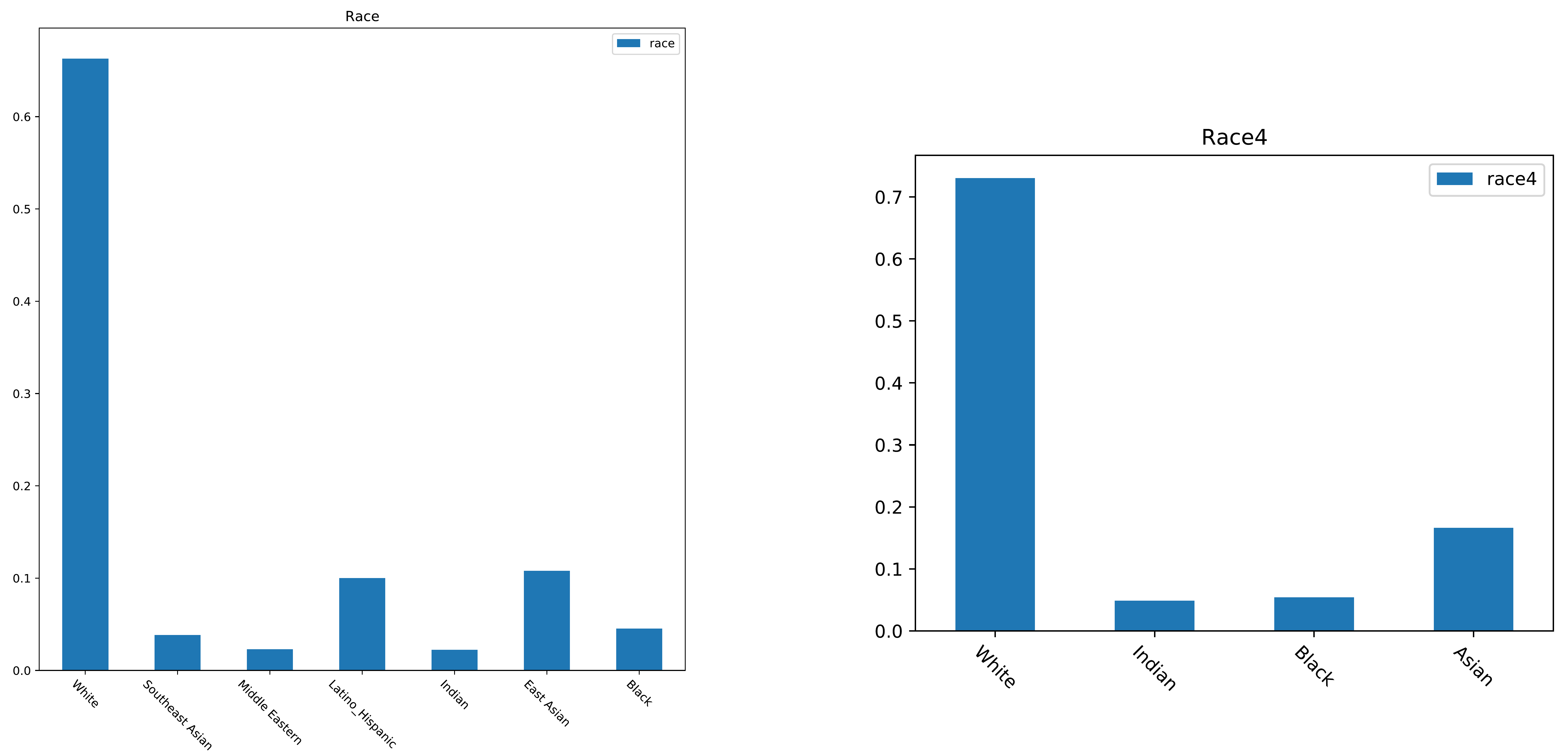}
	\caption{Proportion of faces in each sub-group for Age,Gender,Race and Race4 attribute for GAN generated synthetic faces(x-axis sub-groups and y-axis proportion)}
	\label{fig:fig1}
\end{figure}

\begin{figure}[htp]
	\centering
	\includegraphics[width=.3\textwidth]{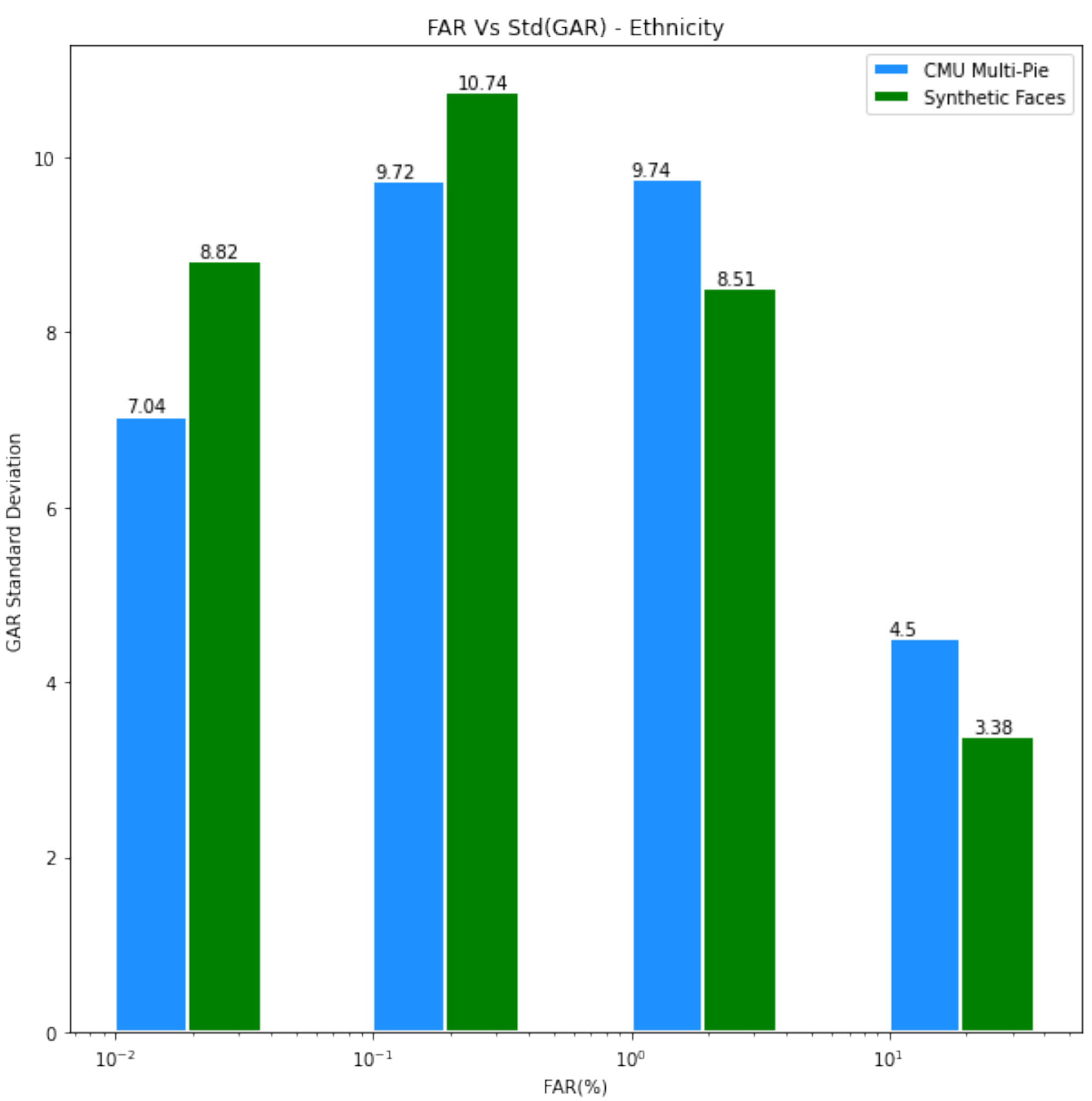}\hspace{1 cm}
	\includegraphics[width=.3\textwidth]{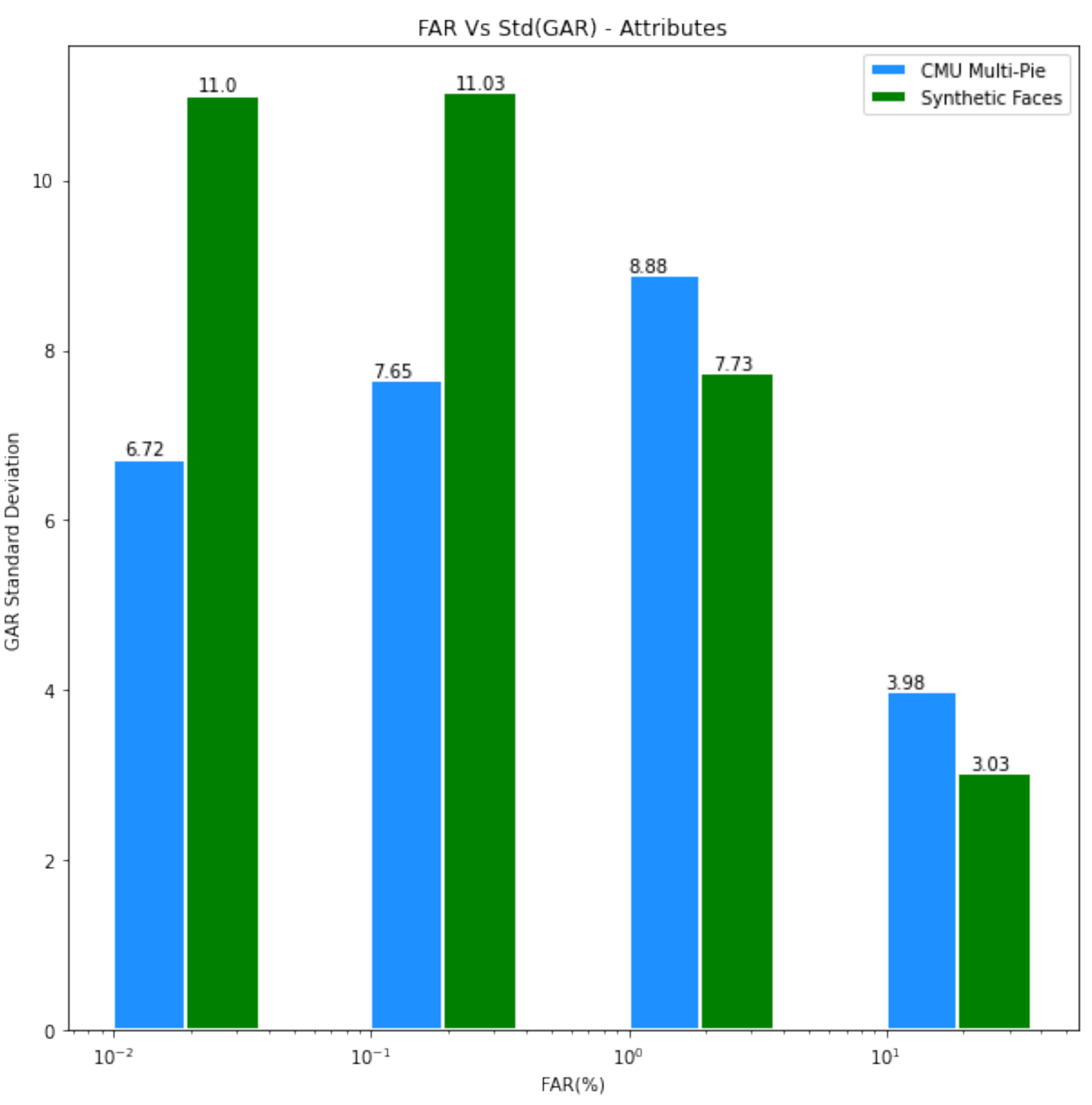}\hfill
	\includegraphics[width=.3\textwidth]{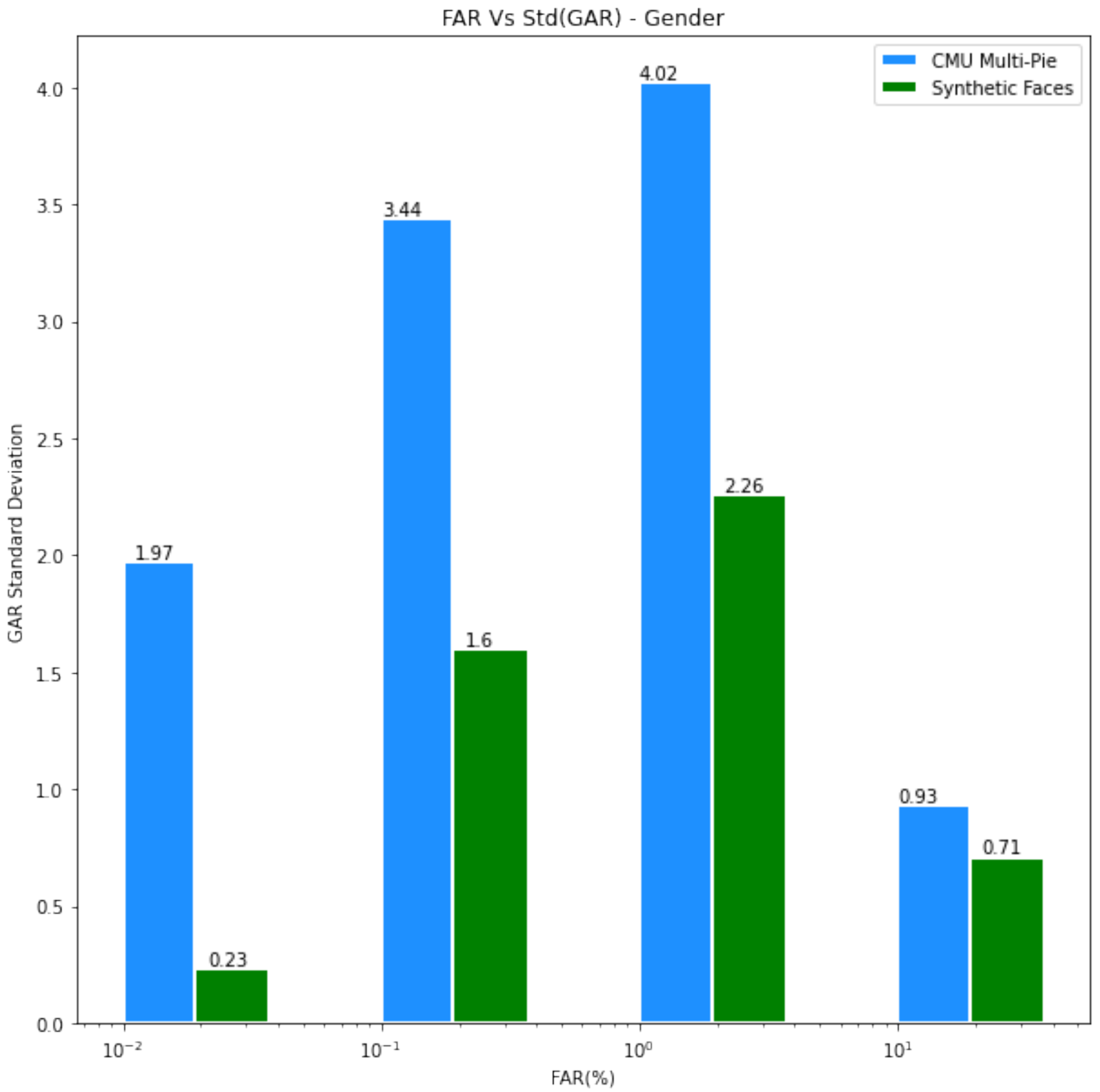}
	\caption{$DoB_{fv}$ i.e Std(GAR @ FAR) for Ethnicity, Gender and Attributes with CMU Multi-Pie and Synthetic faces (smaller is better for bias)}
	\label{fig:fig2}
\end{figure}

\section{Results and Analysis}

From figure \ref{fig:fig1}, it is evident that GANs trained with the FFHQ dataset are biased towards generating more faces in the age group "20-29" and mostly "White" faces. However, no such imbalance is observed for gender attribute.

Figure \ref{fig:fig2}  shows the performance of Face Verification models for different face attributes such as Ethnicity, Attributes and Gender. Bias and fairness is measured by comparing the  $DoB_{fv}$ for models fine-tuned with CMU MultiPie and Synthetic faces. $DoB_{fv}$ is greater for models trained with Synthentic faces. This is predominant at low FAR rates. This behaviour is not observed at high FAR rates. Our observations from the analysis of the results are as follows :
(Observation-1 is drawn from experiment-1 and observations-2,3,4 were drawn from experiment-2)
\let\labelitemi\labelitemii
\begin{itemize}
	\item \textbf{Observation-1:} GANs are biased towards age group "20-29" and "White" faces.
	\item \textbf{Observation-2:} Face Verification models trained or fine-tuned with Synthetic faces exhibit bias for "race" attribute. This is confirmed by high $DoB_{fv}$ for Synthetic faces when compared to CMU MultiPie.
	\item \textbf{Observation-3:} Face Verification models trained or fine-tuned with Synthetic faces doesn't exhibit any bias for "gender" attribute.
	\item \textbf{Observation-4:} At, high FAR rates we don't observe bias (low $DoB_{fv}$). We hypothesize that although biases are present these are masked by high false acceptances.
\end{itemize}

\section{Conclusion}
GANs are popular networks that are very successful in generating faces of good perpetual quality. These are trained with existing datasets. However, the biases present in the dataset are also being manifested in these networks. We analyzed the biases of these networks for important attributes such as age, race and gender for faces. We also demonstrated how this could impact the sub-group performance of downstream models such as face verification systems. Hence, it is important to debiase GANs before using them in any application. In future, we aim to investigate methods and techniques for debiasing GANs with respect to different critical attributes.

\clearpage
%
%
\nocite{*}
\bibliographystyle{splncs04}
\bibliography{egbib}

\clearpage

\appendix

\section{Detailed Description and Visualization of Datasets, Architectures and Results\label{app:experiments_desc}}

\subsection{Datasets}

\begin{figure}[!ht]
	\centering
	\includegraphics[height=4.5cm]{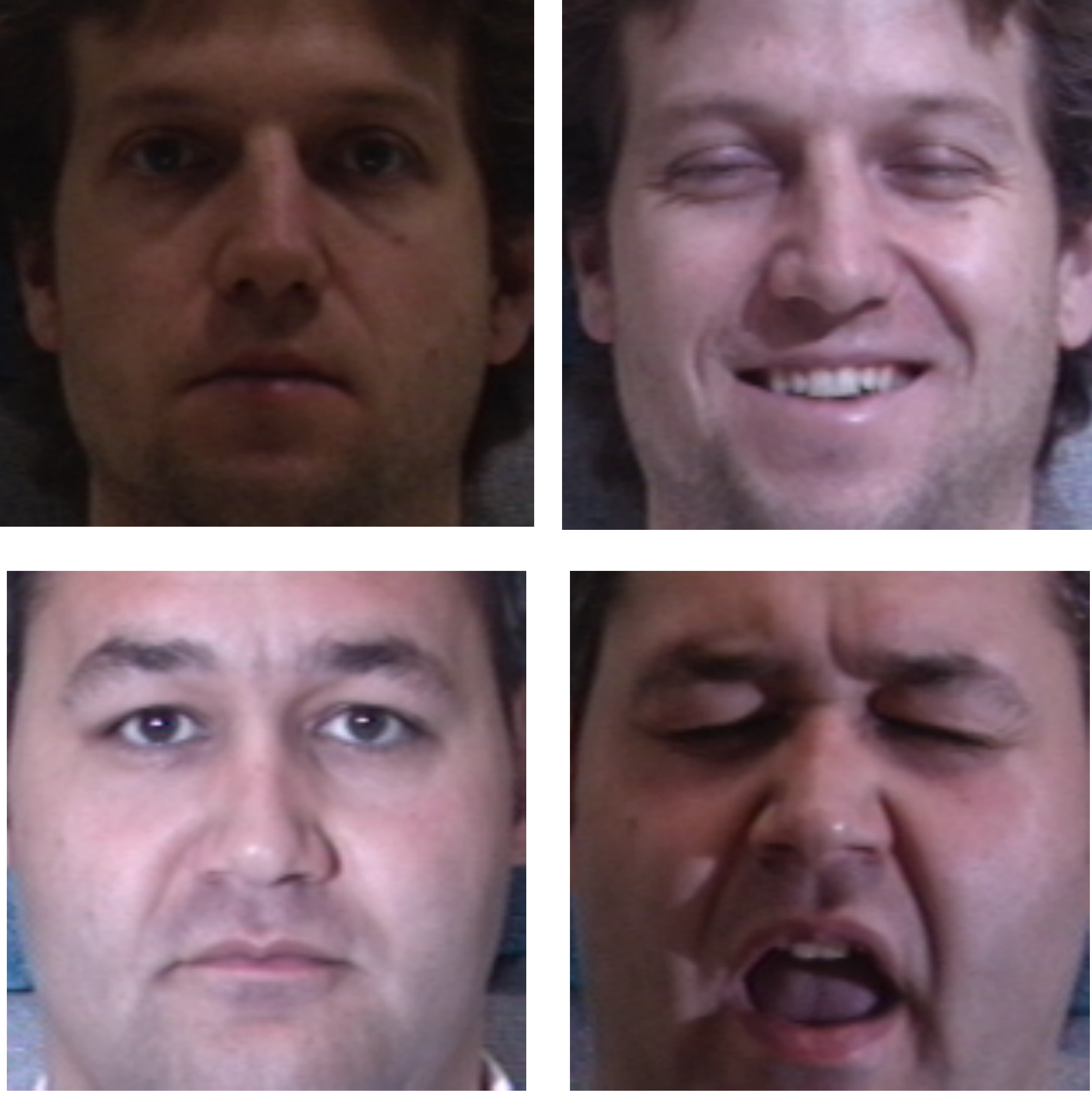}\hspace{1 cm}
	\includegraphics[height=4.5cm]{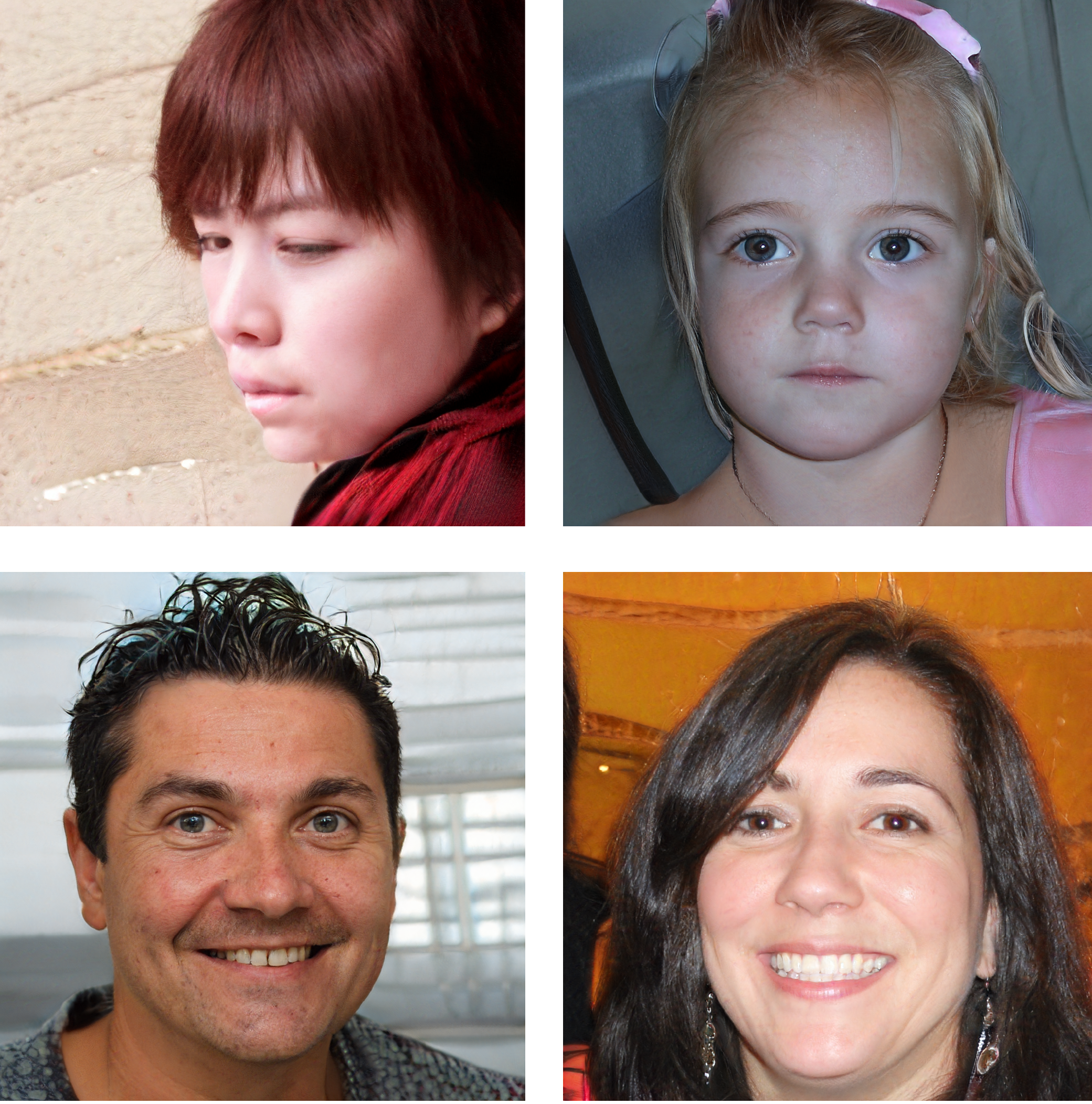}
	\caption{CMU Multi-PIE and FFHQ Datasets}
	\label{fig:fig3}
\end{figure}

Figure \ref{fig:fig3} shows CMU Multi-PIE~\cite{gross2010multi} which is constrained dataset and FFHQ, which stands for Flickr-Faces-HQ~\cite{Karras_2019_CVPR}

\begin{figure}[!ht]
	\centering
	\includegraphics[height=4.5cm]{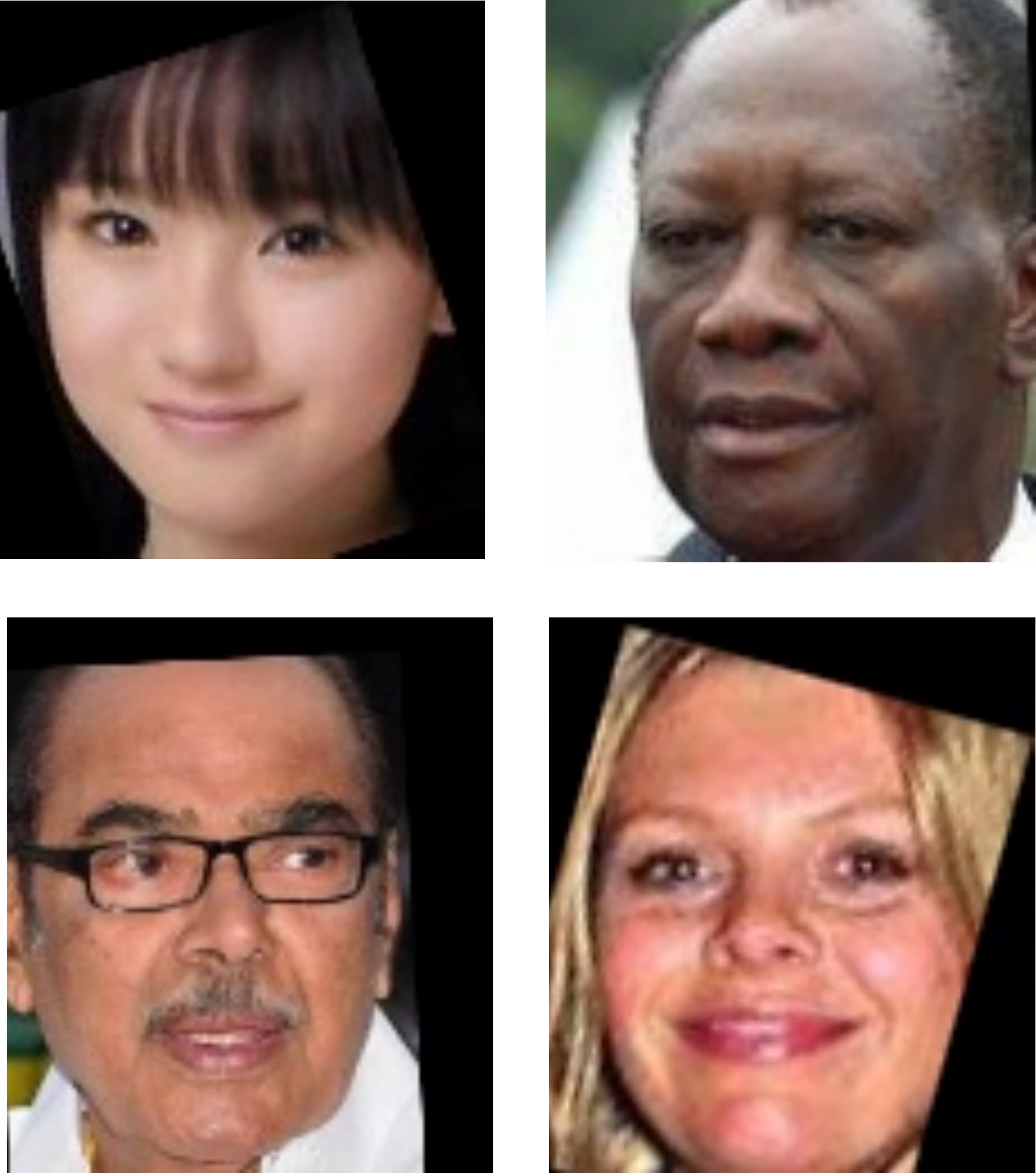}\hspace{0.5 cm}
	\includegraphics[height=4.5cm]{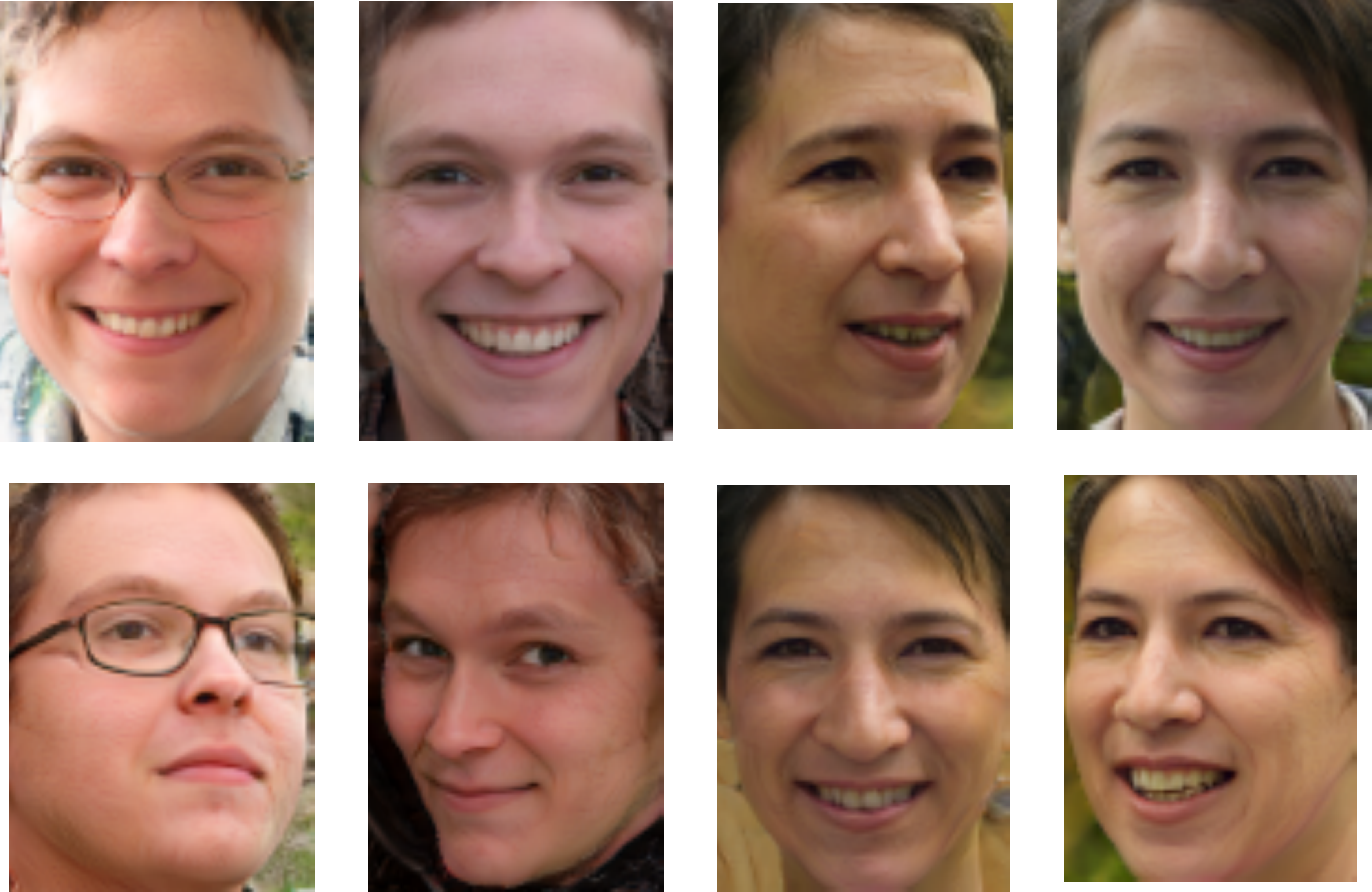}
	\caption{BFW and Synthetic(DiscoFaceGAN) Faces}
	\label{fig:fig4}
\end{figure}

Figure \ref{fig:fig4} shows Balanced Faces in the Wild (BFW)~\cite{robinson2020face} which are used for evaluating bias for face verification task and Synthetic Faces generated with DiscoFaceGAN~\cite{Deng_2020_CVPR}

\subsection{Architectures}

\begin{figure}[!ht]
	\centering
	\includegraphics[width=\textwidth,height=3.0cm]{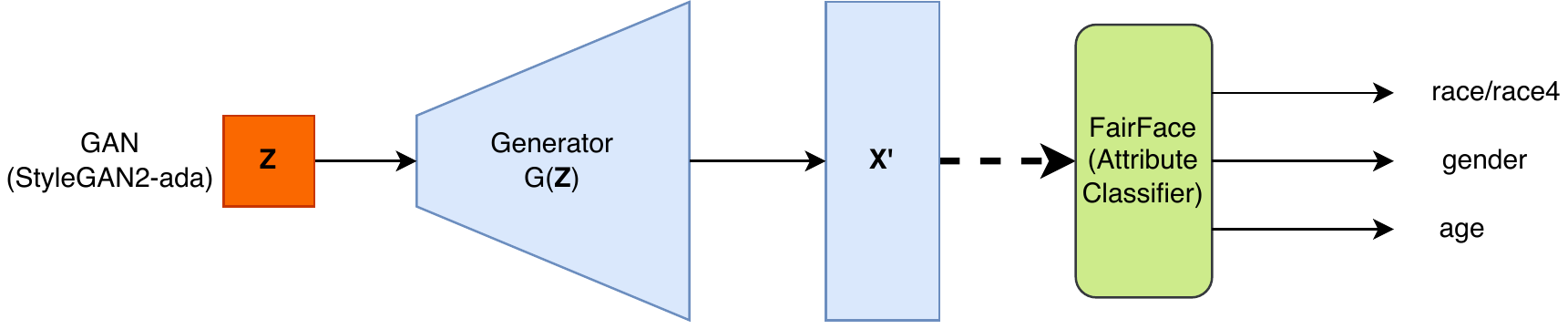}
	\caption{GAN Bias Estimation Architecture}
	\label{fig:fig5}
\end{figure}

Attributes such as race, gender and age for synthetic faces generated by GAN are obtained using a pretrained Fairface ~\cite{karkkainenfairface} attribute classifier. The proportion of images for each attribute are analyzed for imbalance and bias. This architecture is shown in Figure \ref{fig:fig5}

\begin{figure}[!ht]
	\centering
	\includegraphics[height=4.5cm]{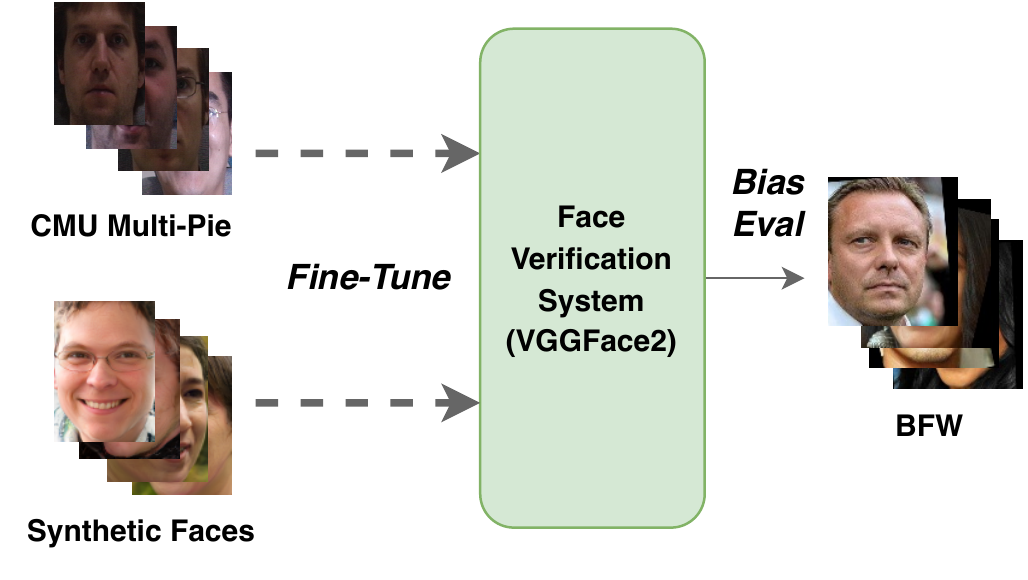}
	\caption{Bias Estimation in Face Verification System}
	\label{fig:fig6}
\end{figure}

As showin in Figure \ref{fig:fig6} the impact of bias and fairness on face verification systems is analyzed by fine-tuning with CMU Multi-PIE~\cite{gross2010multi} and Synthetic Faces generated with DiscoFaceGAN~\cite{Deng_2020_CVPR} and comparing $DoB_{fv}$ i.e Std(GAR @ FAR) for different attributes

\subsection{Results}

Table \ref{tab:mytab1} shows GAR@FAR when face recognition model was fine-tuned with CMU Muti-PIE and Synthetic faces. The overall performance is similar for both the datasets. Table \ref{tab:mytab2}  and Table \ref{tab:mytab3}  shows GAR@FAR and their standard deviations ($DoB_{fv}$) for each sub-group of gender and ethnicity attributes.

\begin{table}[h!]
	\centering
	\begin{tabular}{c c c} 
		\hline
		& \multicolumn{2}{c}{GAR(\%)}  \\
		\hline
		FAR(\%) & CMU Multi-PIE & Synthetic Faces \\
		\hline
		0.01 & 21.59 & 22.77   \\ 
		0.1 & 38.45 & 39.51   \\
		1 & 62.61 & 63.07   \\ 
		10 & 88.02 & 88.05 \\ [1ex] 
		\hline
	\end{tabular}
	\caption{\label{tab:mytab1} GAR@FAR}
\end{table}

\begin{table}[h!]
	\centering
	\begin{tabular}{c c c c c c c} 
		\hline
		& \multicolumn{6}{c}{GAR(\%)}  \\
		\hline
		FAR(\%) &  \multicolumn{3}{c}{CMU Multi-PIE} &  \multicolumn{3}{c}{Synthetic Faces} \\
		\hline
		& Male(M) & Female(F) & Std & Male(M) & Female(F) & Std \\
		\hline
		0.01 & 22.77 & 19.98 & 1.97 & 22.71 & 23.04 & 0.23 \\ 
		0.1 & 41.47 & 36.6 & 3.44 & 40.62 & 38.36 & 1.60 \\
		1 & 66.23 & 60.55 & 4.02 & 64.63 & 61.43  & 2.26 \\ 
		10 & 88.85 & 87.54 & 0.93 & 88.54 & 87.53 & 0.71\\ [1ex] 
		\hline
	\end{tabular}
	\caption{\label{tab:mytab2} GAR@FAR for gender attribute}
\end{table}

\begin{table}[htbp]
	\footnotesize
	\setlength\tabcolsep{-0.2pt}
	\setlength\thickmuskip{0mu}
	\setlength\medmuskip{0mu}
	\hskip-0.4cm
	\begin{tabular}{c c c c c c c c c c c}
		\hline
		& \multicolumn{10}{c}{GAR(\%)}  \\
		\hline
		FAR(\%) &  \multicolumn{5}{c}{CMU Multi-PIE} &  \multicolumn{5}{c}{Synthetic Faces} \\
		\hline
		& Asian(A) & Black(B) & Indian(I) & White(W) &Std & Asian(A) & Black(B) & Indian(I) & White(W) &Std \\
		\hline
		0.01 & 16.2 & 22.24 & 24.37 & 31.19 & 7.04& 18.0 & 18.53 & 27.5 & 36.69 & 8.82 \\ 
		0.1 & 30.05 & 38.27 & 43.57 & 53.23 & 9.72 & 31.35 & 36.75 & 44.48 & 56.11 & 10.74 \\
		1 & 52.46 & 62.25 & 65.52 & 76.08 & 9.74 & 56.55 & 60.08 & 64.95 & 76.10 & 8.51 \\ 
		10 & 82.38 & 88.09 & 87.85 & 93.4 &4.50 & 84.5 & 86.86 & 88.06 & 92.56 & 3.38\\ [1ex] 
		\hline
	\end{tabular}
	\caption{\label{tab:mytab3} GAR@FAR for ethnicity attribute}
\end{table}

\end{document}